\documentclass{article}
\usepackage[T1]{fontenc}
\usepackage{iclr2027_conference,times}
\iclrfinalcopy
\usepackage{amsmath,amssymb,graphicx,booktabs,tabularx,multirow,xspace,microtype}
\usepackage[ruled,vlined]{algorithm2e}
\usepackage{float}
\usepackage{xcolor,subcaption,colortbl,wrapfig}
\definecolor{risarow}{HTML}{EDF3FB}
\newif\ifshowrevision
\showrevisionfalse
\definecolor{revisionblue}{HTML}{1757A1}
\newcommand{\revisioncolor}{\ifshowrevision\color{revisionblue}\else\color{black}\fi}
\DeclareRobustCommand{\revisiontext}[1]{{\revisioncolor#1}}
\usepackage{hyperref}
\hypersetup{hidelinks,pdftitle={RISA: Recoverable Sparse Attention for Block-Diffusion Language Models},pdfauthor={Xinrui Zhong, Fanjiang Ye, Yuke Wang}}
\usepackage[capitalize,noabbrev]{cleveref}
\usepackage{xurl,wrapfig}
\graphicspath{{figures/}}
\DeclareRobustCommand{\method}{\textup{RISA}\xspace}
\newcommand{\hot}{\textsc{Hot}\xspace}
\newcommand{\cold}{\textsc{Cold}\xspace}

\title{\textup{RISA}: Recoverable Sparse Attention for Block-Diffusion Language Models}
\author{\makebox[\textwidth][c]{\bfseries Xinrui Zhong\quad Fanjiang Ye\quad Yuke Wang}\\
\makebox[\textwidth][c]{\normalfont Rice University}}
\begin{document}
\maketitle
\begin{abstract}
Block-wise diffusion language models (DLMs) refine multiple tokens in parallel, offering an alternative to token-by-token autoregressive decoding. Their long-context inference, however, remains bottlenecked by memory-bound attention: every denoising step repeatedly reads a growing historical key--value (KV) cache. Directly applying sparse attention does not fully solve this problem, because different queries can select different pages, inflating the block's collective read set, while repeated support construction adds overhead and permanent eviction can discard information needed later. We present \method, a recoverable sparse-attention method with shared read plans for block-wise DLMs. \method exploits the fact that historical KV remains fixed within a block while queries evolve across denoising steps. Newly generated history is organized into exact priority and recovery tiers; seed queries then build a protected, bounded read plan whose indices are reused across the block while attention is recomputed with fresh queries and current-block KV. This separates history preservation from repeated reading and amortizes sparse-support construction without changing the backbone or decoder. On DreamReasoner-8B, \method retains comparable task performance. \revisiontext{At 32K, the default configuration speeds up attention by $2.67\times$ and fixed-trajectory requests by $1.36\times$ over dense.} On H200, attention speedup reaches $5.53\times$ at 128K. \revisiontext{Across eight settings on three models, \method decodes faster than the tested MAGE, QUEST, and LoSA budgets at matched KL.}
\end{abstract}
\section{Introduction}
\label{sec:intro}

Diffusion language models (DLMs) generate text by iteratively refining multiple token positions in parallel, offering an alternative to token-by-token autoregressive decoding \citep{austin2021d3pm,lou2024sedd,sahoo2024mdlm}. Large-scale models such as LLaDA and Dream demonstrate the potential of this paradigm for language modeling, instruction following, and reasoning \citep{nie2025llada,ye2025dream}. Block-wise DLMs combine parallel refinement within each block with autoregressive progression across blocks, allowing completed blocks to serve as cached context for subsequent generation \citep{arriola2025blockdiffusion,wu2026dreamreasoner}. This parallelism, however, does not remove the cost of accessing history. Caching avoids \emph{recomputing} committed key--value (KV) states, but every denoising step still \emph{reads} the growing prefix to compute attention. For long prompts and extended reasoning traces, these repeated KV accesses become increasingly memory-bound and limit the decoding speedup that block-wise parallelism can deliver \citep{kwon2026mage,xi2026losa}.

Sparse attention is a natural way to reduce this cost by reading only a subset of the historical KV \citep{zhang2023h2o,tang2024quest,xiao2024infllm}. Yet sparsity designed for ordinary autoregressive decoding does not directly translate into efficient block diffusion. A block contains many evolving queries, and independently selecting a small support for each query can produce a much larger union of accessed pages---the \emph{KV inflation} problem identified by \citet{xi2026losa}. Reconstructing these supports across denoising steps further introduces repeated selection overhead. An alternative is to permanently evict low-priority KV, but doing so can remove information that later blocks need. Recent diffusion-specific methods therefore amortize attention computation through pattern, index, or output reuse \citep{wang2026sparsed,kwon2026mage,xi2026losa}. The resulting design challenge is to bound repeated KV reads without permanently discarding historical information that later blocks may still need under different queries.

Block diffusion provides a useful structure for addressing this question. During the denoising of one block, historical KV remains fixed while the queries associated with the unfinished block continue to evolve. This creates a natural block-level scope for selecting historical addresses once and reusing them across subsequent denoising steps, while still recomputing attention with fresh queries. At the same time, the repeated queries issued before a generated block ages out provide multiple observations of how its tokens are used. These observations can inform longer-term priority without forcing an irreversible decision immediately after generation, giving the block a natural scope for read selection and history management without assuming static attention.

\begin{figure}[t]
\centering
\includegraphics[width=\linewidth]{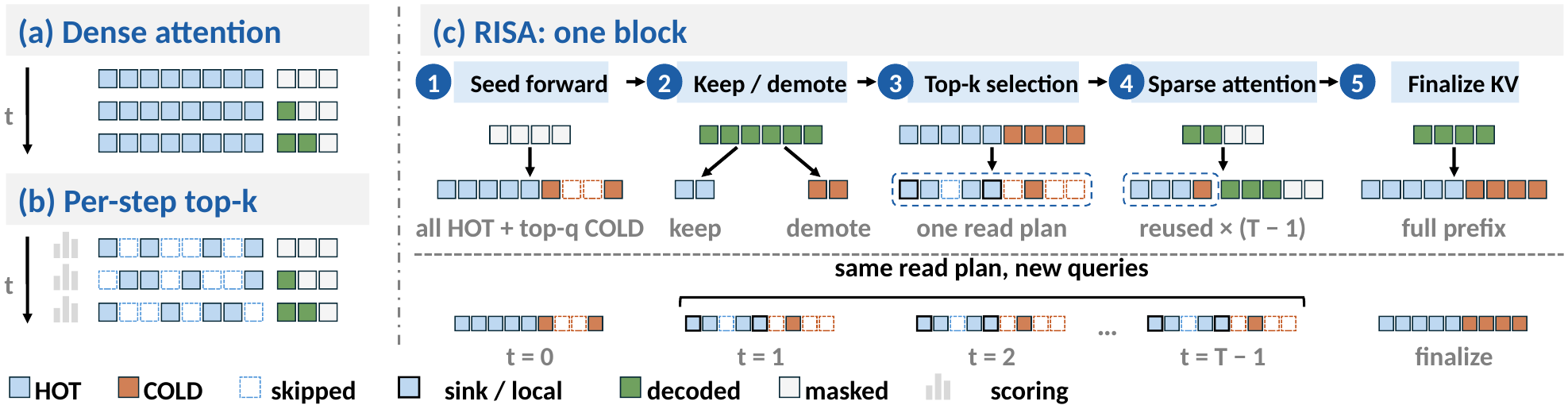}
\caption{\textbf{Dense rereads history; \method reuses one bounded read plan.}
(a) Dense attention revisits the historical prefix at every denoising step. (b) Per-step top-$k$ reduces each attention support but repeats scoring and selection. (c) \method uses the seed forward to observe available history, admits aging KV, constructs one protected read plan, and reuses its historical indices for later steps while queries continue to evolve; finalization still reads the full legal prefix.}
\label{fig:overview}
\end{figure}

We introduce \method, a sparse-attention method that separates \emph{preserving historical KV} from \emph{deciding which history a block reads} (\cref{fig:overview}). Newly generated KV first remains in a grace window, where repeated denoising queries provide observations before it is assigned to a compact priority tier or an addressable recovery tier; both retain exact KV values on the GPU. For each new block, seed queries jointly select a protected, bounded page plan across the available history. Subsequent denoising steps reuse these indices while recomputing attention with their current queries and current-block KV, and later blocks may recover history omitted from an earlier plan. The backbone, native decoder, commit rule, and full-prefix KV finalization remain unchanged throughout generation.

Our main contributions are:
\begin{itemize}
    \item We introduce a \textbf{recoverable history organization} that uses repeated denoising observations to prioritize generated KV while keeping lower-priority history available for later blocks.
    \item We develop a \textbf{block-static sparse-attention executor} that bounds older-history reads and reuses one page plan across denoising steps while recomputing attention with fresh queries.
    \item We provide a \textbf{cross-model quality and system evaluation} across reasoning, code, retrieval, and long-context settings. On DreamReasoner-8B, \method retains comparable task performance. \revisiontext{At 32K, the default configuration speeds up attention by $2.67\times$ and fixed-trajectory requests by $1.36\times$.} At 128K on H200, it reaches $5.53\times$ attention speedup. \revisiontext{Across the eight in-window settings, it decodes faster than the tested MAGE, QUEST, and LoSA budgets at matched KL.}
\end{itemize}

\section{Related Work}
\label{sec:related_work}
\paragraph{Diffusion language models.}
Discrete and masked diffusion generate text by reversing token corruption \citep{austin2021d3pm,lou2024sedd,sahoo2024mdlm}; LLaDA and Dream scale this approach to language modeling \citep{nie2025llada,ye2025dream}. Block Diffusion combines within-block denoising with autoregressive progression and prefix caching \citep{arriola2025blockdiffusion}, while DreamReasoner develops block-size curricula for reasoning \citep{wu2026dreamreasoner}. We target the repeated attention reads that remain after historical representations have been cached.

\paragraph{KV compression and sparse attention.}
FlashAttention optimizes exact attention through IO-aware execution and work partitioning \citep{dao2022flashattention,dao2024flashattention2}. KV compression uses observed attention \citep{zhang2023h2o,liu2023scissorhands,li2024snapkv}, sinks and recency \citep{xiao2024streamingllm}, or head specialization \citep{xiao2025duoattention}. Retrieval methods preserve a larger history and select query-relevant KV \citep{tang2024quest,ribar2024sparq,xiao2024infllm}. TidalDecode reuses indices across layers \citep{yang2025tidaldecode}; HiSparse combines host storage with bounded GPU residency \citep{xie2026hisparse}. \method instead keeps both tiers on the GPU: priority determines access preference, not deletion, and the objective is reduced read traffic rather than KV-memory compression.

\paragraph{Efficient diffusion inference.}
Fast-dLLM introduces approximate block-wise KV caching for bidirectional DLMs and confidence-aware parallel decoding \citep{wu2026fastdllm}; dKV-Cache likewise targets redundant representation computation \citep{ma2025dkvcache}. Caching reduces what must be recomputed, while parallel decoding changes which tokens are committed together. Cached KV can nevertheless be read in full at every step. \method instead reduces these reads through sparse attention while preserving the backbone and native commit and stopping rules. It addresses a complementary bottleneck rather than replacing the cache or decoder.

Within sparse diffusion inference, SparseD reuses head-specific patterns \citep{wang2026sparsed}, MAGE derives token indices from exact all-mask attention and reuses them within a block \citep{kwon2026mage}, and Sparse-dLLM combines eviction with block-boundary refresh \citep{song2025sparsedllm}. LoSA instead reuses prefix-attention outputs for stable tokens \citep{xi2026losa}. \method recomputes attention for every current-block query and selects jointly from page summaries. Its distinction is not index reuse alone, but combining a protected, bounded read plan with observation-guided, recoverable generated history that remains addressable across later blocks.

\section{Motivation: Preserve History, Avoid Repeated Work}
\label{sec:motivation}
Sparse attention is useful only when the cost it removes is not replaced by irreversible cache decisions or repeated support construction. We therefore ask two questions: should low-priority history remain recoverable, and must a block rebuild its support at every denoising step? The two decisions operate at different time scales: storage priority persists across blocks, while a read plan serves only the block currently being refined. Diagnostic alternatives reuse the same sampled queries, KV values, and masks, so they isolate these mechanisms without changing the generated trajectory; protocol details appear in the appendix.

\subsection{Low Priority Does Not Mean No Future Use}
\label{sec:recovery_motivation}
Attention-based eviction estimates importance from observations available so far \citep{zhang2023h2o,li2024snapkv}. Such evidence can prioritize what should be read frequently, but it is weaker evidence for permanent deletion because later blocks issue different queries. We therefore treat priority as a scheduling decision rather than a claim that omitted history has become irrelevant.

\begin{figure}[t]
\centering
\begin{subfigure}[t]{.48\linewidth}\centering
\includegraphics[width=\linewidth]{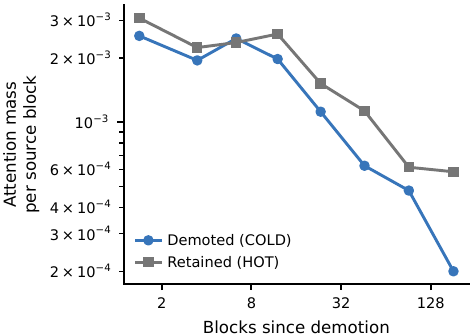}\caption{}\end{subfigure}\hfill
\begin{subfigure}[t]{.48\linewidth}\centering
\includegraphics[width=\linewidth]{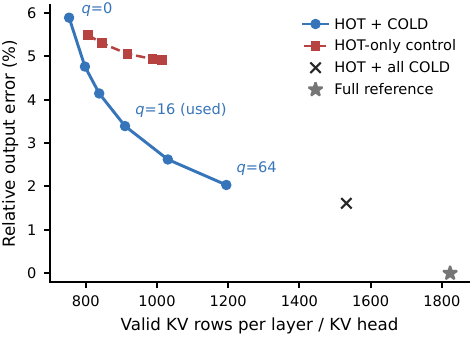}\caption{}\end{subfigure}
\caption{\textbf{Low-priority history can remain useful.} (a) Full-domain attention continues to revisit tokens after they have been demoted from the priority tier. (b) At a comparable read budget, retrieving from the recovery tier yields lower attention-output error than spending the same budget only on additional priority pages. Error bars in (a) resample requests.}
\label{fig:motivation_recovery}
\end{figure}

\Cref{fig:motivation_recovery}a shows that demoted cohorts continue to receive attention from later blocks. Moreover, redirecting the same extra read budget to additional HOT pages leaves a larger approximation error than retrieving COLD pages (\cref{fig:motivation_recovery}b). Recent use is therefore informative for prioritization, but not a reliable certificate for irreversible eviction; recoverable storage preserves this distinction.

\subsection{Bound Reads and Amortize Selection}
\label{sec:selection_motivation}
Query-aware retrieval reduces KV reads but introduces scoring, ranking, and dispatch \citep{tang2024quest}. In block diffusion these costs can repeat even though historical KV is unchanged: attention weights evolve, but one historical address set can often serve several denoising queries.

\begin{figure}[t]
\centering
\includegraphics[width=\linewidth]{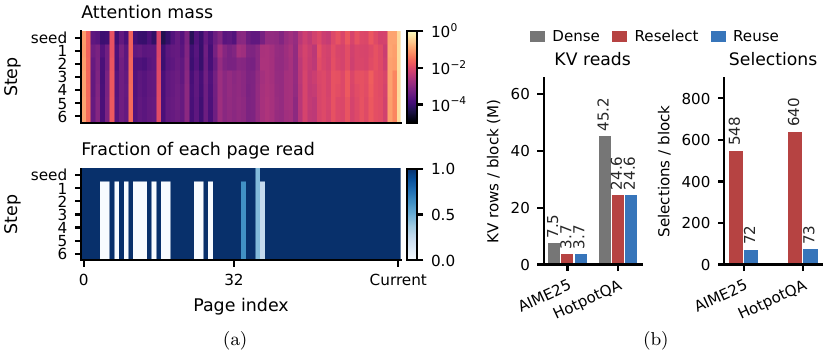}
\caption{\textbf{Read sparsity and index reuse remove different work.} (a) Attention mass changes across denoising, while a block-static execution can keep the same historical support. (b) Sparse execution reduces requested KV rows, and reusing the plan avoids repeatedly rebuilding that support. Counts include seed and mandatory work and should not be interpreted as direct HBM-traffic measurements.}
\label{fig:mass}
\end{figure}

\Cref{fig:mass} illustrates the two savings. Sparse support bounds the historical rows consumed by attention, whereas plan reuse removes repeated selection without reusing stale attention outputs. The controlled ablation in \cref{sec:ablations} shows the same separation in latency. Sparsity and reuse therefore address complementary costs: what is read, and how often those reads must be selected.

Reuse also requires a bounded read set: a shared plan does not scale if it retains a fixed fraction of an ever-growing prefix. We therefore protect recent or otherwise mandatory context separately and cap eligible older HOT reads. This separates two complementary sources of savings within the same block: sparse support limits recurring KV reads, while plan reuse amortizes the work of finding that support. The main ablation confirms that the cap lowers attention cost \revisiontext{where it binds (\cref{tab:main_ablation})}. The resulting plan bounds recurring reads while leaving lower-priority history recoverable for later blocks and longer prefixes, rather than turning a read-priority decision into permanent eviction.

\section{\texorpdfstring{\textup{RISA}}{RISA}}
\label{sec:method}
\begin{figure}[t]
\centering
\includegraphics[width=\linewidth]{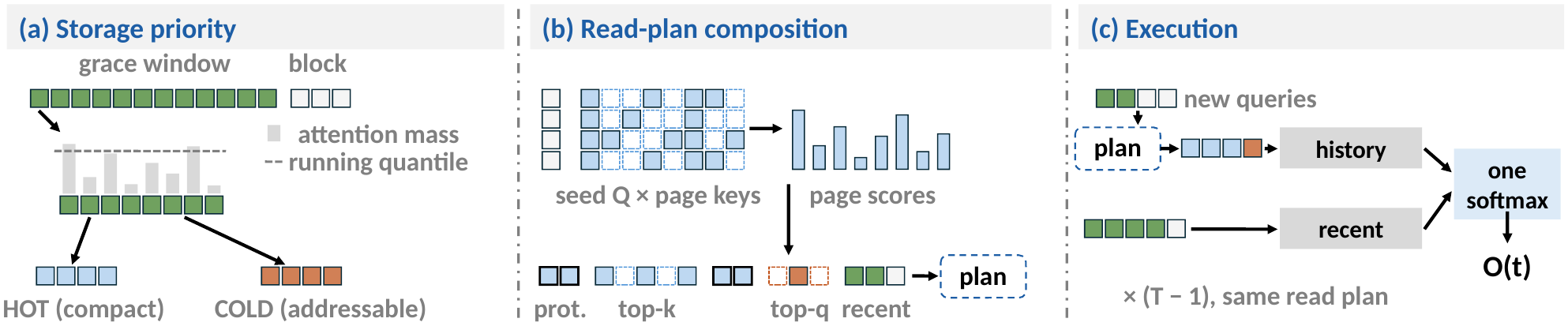}
\caption{\textbf{Storage priority, read-plan composition, and execution.} (a) Grace-window observations determine whether generated KV enters compact \hot or addressable \cold storage; both keep exact values. (b) Seed queries score page summaries, then protected HOT, selected HOT, recovered COLD, and mandatory recent rows form one read plan. (c) Later steps reuse the historical plan with new queries and current recent KV, combining all contributions under one softmax.}
\label{fig:method}\label{fig:hierarchy}\label{fig:execution}
\end{figure}

\method separates the lifetime of stored history from the lifetime of its read plan (\cref{fig:method}). The three panels expose this separation directly: observations determine storage priority as blocks age, seed queries compose a bounded plan, and later steps reuse only its historical addresses while recomputing attention. Thus, storage state persists across blocks, whereas the sparse plan is temporary state for the block being denoised. Each model uses one fixed budget, with no change to its backbone or decoder.

\subsection{Observe Recent History Before Prioritizing It}
\label{sec:storage}
Newly committed blocks enter a fully readable \emph{grace window}. Repeated queries supply observations before aging entries enter compact \hot storage or addressable \cold storage. Both preserve their stored KV values on the GPU; prompt KV stays in HOT but remains subject to read selection. Future plans can read COLD entries directly without promotion or deletion.

For each token and KV head, a fixed priority rule maps accumulated attention observations to a priority score. A running quantile of prior scores targets eight retained HOT entries per 32-token source block. Observations come from the executed attention distribution, with no additional teacher forward. This score controls long-term storage priority rather than the current block's read set. \revisiontext{We score each token by its grace-window attention mass (\cref{eq:priority}). Mass-only admission closely matches the learned correction on AIME24 (\cref{app:admission}).}

With 32-token blocks, 16-slot pages, and $G=8$, the seed observes the outgoing eighth block before admission. Later steps read the seven more recent blocks, when present, and the active block. Short prefixes below 512 tokens retain the carrier's original dense-attention path.

\paragraph{Storage state versus read state.}
HOT/COLD assignment persists across blocks and determines how historical KV is organized, whereas the read plan is temporary state for the block currently being denoised. A page omitted from one block's plan is therefore not deleted: it remains addressable and can be selected again by a later block. Conversely, retaining a page in storage does not force every step to read it. This separation bounds recurring reads without turning sparsification into irreversible KV eviction, so omitted pages remain eligible for future block plans.

\subsection{Select One Protected, Bounded Plan}
\label{sec:plan}
The seed reads full HOT, mandatory rows, and up to \revisiontext{$q=128$} query-selected COLD pages, producing logits and saved queries. HOT eligibility and protection are prepared before this forward. After observations are updated and aging entries are admitted, the native rule commits seed proposals. If denoising continues, saved queries build the reuse plan. Newly admitted HOT rows remain mandatory for this block and join the ranked candidate set when the next block begins.

Each layer and KV head has one plan shared by its positions and mapped query heads. For each query, we take the larger dot product with a page's minimum or maximum key vector, then sum these scores over the block before selecting pages. This yields one bounded support rather than a union of per-query sets. The max-after-dot heuristic uses page-wise key extrema \citep{tang2024quest}, but differs from QUEST's coordinate-wise bound in how relevance is aggregated across the page.

Let $H_b^{\rm prot}$ contain the first four and last eight sealed HOT pages, protecting initial and recent resident context \citep{xiao2024streamingllm}. From the remaining eligible pages $E_b$, we select a fraction $\rho$, subject to a 512-page cap; $\rho=0.5$ for DreamReasoner \revisiontext{and LLaDA2.2-mini} and $1.0$ for SDAR. The budget $k_b$ rounds $\rho|E_b|$, clips it to the cap and candidate count, and selects at least one page when available. Using pooled page scores $u$, we combine mandatory and selected rows into the block's read support:
\begin{equation}
 S_b=R_b\cup\operatorname{rows}\!\left(H_b^{\rm prot}\cup
 \operatorname{Top}_{k_b}(u,E_b)\cup\operatorname{Top}_{q}(u,C_b)\right),
 \label{eq:support}
\end{equation}
where $C_b$ contains COLD candidates and $R_b$ contains mandatory recent/current, partial-page, and newly admitted HOT rows. The cap excludes protected pages, COLD, and mandatory rows. Each valid row appears once; unselected HOT pages stay available for subsequent block plans.

\subsection{Reuse Addresses, Not Attention Outputs}
\label{sec:execution}
Later steps reuse $S_b$ with fresh queries, current-block KV, and attention outputs. Historical and recent/current partials share one softmax normalization; they are not independently normalized outputs that are averaged. Persistent device buffers hold indices and valid lengths for CUDA-graph replay; a block completed by its seed skips plan construction entirely.

\paragraph{Reuse addresses, recompute attention.}
The shared plan fixes only which historical rows are gathered. Each denoising step still recomputes query--key scores, normalization, and value aggregation using its current queries and current-block state. Thus the optimization does not assume that attention probabilities are static across denoising; it amortizes support construction while preserving step-specific attention computation. Mandatory recent/current rows remain outside the eligible-HOT cap so that newly evolving context is not frozen by the seed plan.

After commitment, the native full-prefix finalization pass reads all legal KV to write the block's representations, independently of $S_b$. Seed attention, maintenance, and finalization therefore remain context-dependent costs. Recent protection and later recovery mitigate omitted-support error without making attention lossless. Exact storage does not imply dense-equivalent trajectories.

\section{Evaluation}
\label{sec:evaluation}
\paragraph{Experimental setup.}
\label{sec:setup}
We evaluate DreamReasoner-8B \citep{wu2026dreamreasoner}, SDAR-8B-Chat \citep{cheng2025sdar}\revisiontext{, and LLaDA2.2-mini \citep{llada22mini}, a mixture-of-experts model with a native 128K window,} in BF16 on NVIDIA H200 GPUs, using 32-token blocks and one request per device. \method uses $G=8$, $K_{\max}=512$, and \revisiontext{$q=128$}, with a fixed read fraction per model: $\rho=0.5$ on DreamReasoner \revisiontext{and LLaDA2.2-mini} and $1.0$ on SDAR. Efficiency is measured at 16K, 32K, 64K, and 128K prefixes. \revisiontext{DreamReasoner and SDAR} are configured for contexts up to 32K; we use the longer prefixes to characterize long-context scalability, while task-quality evaluation remains within the configured range. \revisiontext{LLaDA2.2-mini covers all four contexts natively.}

\paragraph{Metrics.}
\revisiontext{Besides task accuracy, we measure fidelity as the KL divergence from dense on dense's own denoising states,} and two latencies: \emph{attention latency}, the attention path per denoising step including selection and amortized per-block work, and \revisiontext{\emph{per-request latency}, the wall time of a complete decode. Per-request budget sweeps replay dense's trajectory for 512 output tokens, so every method runs the same denoising steps (\cref{app:protocol}).}

\paragraph{Baselines.}
Dense uses each carrier's native full-attention path. QUEST performs query-dependent sparse retrieval, while MAGE forms a block-level sparse pattern from its all-mask observation and reuses that pattern during later denoising steps. As no official MAGE implementation was publicly available at the time of our evaluation, we reimplement it following the algorithm and configuration described in the paper \citep{kwon2026mage}\revisiontext{, overlapping its selection with the forward pass on a separate CUDA stream and keeping its first two layers exact. LoSA \citep{xi2026losa} recomputes only the queries that changed most and caches the rest; we likewise implement it from the paper}. Dense, QUEST, MAGE, \revisiontext{LoSA,} and \method share the same model weights, tokenizer, prompts, native decoder, and block size\revisiontext{, paged attention kernel, and CUDA graphs}. QUEST and MAGE use nominal 1,024-token budgets in the task benchmarks. \method's cap applies to eligible HOT pages, with protected HOT, recovered COLD, and mandatory recent/current rows tracked separately. The budget sweeps record realized reads and compare latency at matched fidelity.

\subsection{End-to-End Accuracy}
\label{sec:accuracy}
\begin{table}[t]
\centering\small
\caption{\textbf{Benchmark accuracy} with 32-token blocks (original configuration). Mathematics uses answer accuracy, LCB execution pass@1, and HotpotQA token F1; bold marks the best sparse method.}
\label{tab:accuracy}\label{tab:sdar8b_accuracy}
\setlength{\tabcolsep}{3.2pt}
\begin{tabular*}{\linewidth}{@{\extracolsep{\fill}}lrrrrrrrr@{}}\toprule
 & \multicolumn{5}{c}{DreamReasoner-8B} & \multicolumn{3}{c}{SDAR-8B-Chat}\\\cmidrule(lr){2-6}\cmidrule(l){7-9}
Method & GSM8K & MATH-500 & AIME25 & LCB & HotpotQA & GSM8K & MATH-500 & HotpotQA\\\midrule
Dense & 85.0 & 88.2 & 58.3 & 50.0 & 40.2 & 88.5 & 65.4 & 53.6 \\
\midrule
QUEST & \textbf{86.0} & 88.2 & 59.4 & 47.5 & 38.9 & 87.5 & 66.2 & 52.3 \\
MAGE & 85.5 & \textbf{89.0} & 59.2 & 42.5 & 35.2 & 88.0 & 66.2 & 51.4 \\
\method & \textbf{86.0} & 88.8 & \textbf{60.4} & \textbf{50.0} & \textbf{39.9} & \textbf{89.5} & \textbf{66.8} & \textbf{53.4} \\
\bottomrule
\end{tabular*}
\end{table}

\paragraph{Benchmarks.}
DreamReasoner-8B is evaluated on GSM8K, MATH-500, AIME25, LiveCodeBench v6, and HotpotQA; SDAR-8B-Chat is evaluated on GSM8K, MATH-500, and HotpotQA. All methods use identical prompts, decoding settings, and generation limits within each model and benchmark. Mathematics uses answer accuracy, LiveCodeBench uses execution pass@1, and HotpotQA uses token F1. GSM8K, MATH-500, LiveCodeBench, and HotpotQA use greedy decoding; AIME25 uses temperature $0.6$ with prompt and output together limited to 32,768 positions. Output caps are 24,576 tokens for mathematics/code and 1,024 for HotpotQA. \revisiontext{Benchmark scores and the DreamReasoner and SDAR attention profiles (\cref{fig:efficiency}) use $q=16$, the learned admission correction, and $\rho=0.75$ on SDAR.}

\paragraph{DreamReasoner-8B.}
\Cref{tab:accuracy} reports competitive scores across reasoning, code, and retrieval. \method obtains $86.0\%$ on GSM8K and $88.8\%$ on MATH-500, compared with dense's $85.0\%$ and $88.2\%$. Its $50.0\%$ LiveCodeBench accuracy matches dense and exceeds QUEST and MAGE by $2.5$ and $7.5$ points. On HotpotQA, it is $0.3$ F1 below dense but $1.0$ and $4.7$ above the sparse alternatives. The ranking is task-dependent: \method ties QUEST on GSM8K, while MAGE is $0.2$ points higher on MATH-500. On AIME25, \method scores $60.4\%$, versus $58.3\%$ for dense, $59.4\%$ for QUEST, and $59.2\%$ for MAGE. Overall, \method remains close to dense while matching or exceeding the reported sparse alternatives on four of the five DreamReasoner tasks.

\noindent\textbf{SDAR-8B-Chat.}
We use the same history-management design on SDAR with the fixed configuration stated above (\cref{tab:accuracy}). \method reaches $89.5\%$ on GSM8K and $66.8\%$ on MATH-500, versus dense's $88.5\%$ and $65.4\%$, and is $0.2$ F1 below dense on HotpotQA. It leads the sparse baselines on all three tasks under identical prompts, benchmark settings, and generation limits.

\subsection{Efficiency Evaluation}
\label{sec:tradeoff}\label{sec:breakdown}\label{sec:efficiency_comparison}\label{sec:component_latency}
\begin{figure}[t]
\centering
\includegraphics[width=\linewidth]{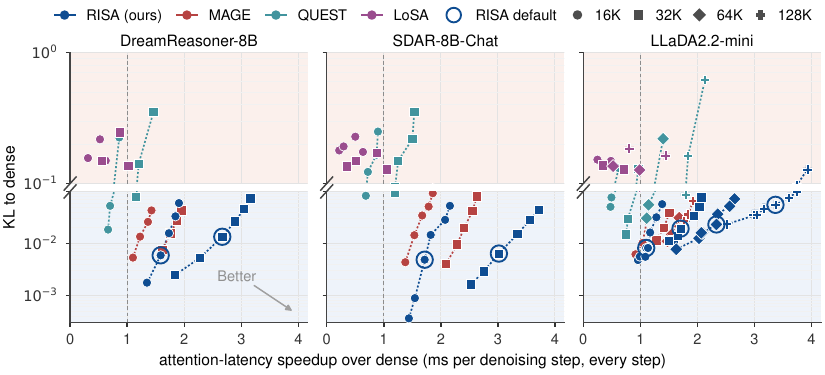}
\caption{\revisiontext{\textbf{Quality--speed frontiers.} Attention-latency speedup over dense (ms per denoising step on natively decoded 512-token requests, per-block work spread over dense's steps per block) against KL to dense on dense's states; lower right is better. Dashed line: dense ($1\times$). The two y regions use separate log scales. Color: method; shape: context; dotted lines: each method's frontier; ring: \method's default; no point is clipped. At LLaDA2.2-mini 16K, MAGE's 8K and \method's $\rho=1.0$ budgets are slightly slower than dense.}}
\label{fig:frontier}
\end{figure}

\begin{table}[t]
\centering\scriptsize
\caption{\revisiontext{\textbf{Per-request speedup over dense} (decode wall time; dense $=1.00\times$) for complete 512-token requests, with every method replaying dense's denoising trajectory on the same documents; \method's default, MAGE and QUEST at 8,192 tokens, LoSA at 4,096. $^\dagger$Beyond DreamReasoner's and SDAR's configured 32K window (YaRN); execution only.}}
\label{tab:request}
\setlength{\tabcolsep}{2.6pt}
{\revisioncolor
\begin{tabular*}{\linewidth}{@{\extracolsep{\fill}}lrrrrrrrrrrrr@{}}\toprule
 & \multicolumn{4}{c}{DreamReasoner-8B} & \multicolumn{4}{c}{SDAR-8B-Chat} & \multicolumn{4}{c}{LLaDA2.2-mini}\\\cmidrule(lr){2-5}\cmidrule(lr){6-9}\cmidrule(l){10-13}
Method & 16K & 32K & 64K$^\dagger$ & 128K$^\dagger$ & 16K & 32K & 64K$^\dagger$ & 128K$^\dagger$ & 16K & 32K & 64K & 128K\\\midrule
\method & \textbf{1.15} & \textbf{1.36} & \textbf{1.76} & \textbf{2.67} & \textbf{1.17} & \textbf{1.40} & \textbf{1.76} & \textbf{2.67} & \textbf{1.07} & \textbf{1.06} & \textbf{1.16} & \textbf{1.25} \\
MAGE 8K & 0.99 & 1.23 & 1.43 & 2.03 & 1.07 & 1.30 & 1.47 & 2.07 & 0.95 & 1.03 & 1.00 & 1.14 \\
QUEST 8K & 0.92 & 1.08 & 1.35 & 2.04 & 0.91 & 1.11 & 1.36 & 2.02 & 1.01 & 1.00 & 1.10 & 1.19 \\
LoSA 4K & 0.84 & 1.01 & 1.18 & 1.64 & 0.86 & 1.03 & 1.19 & 1.64 & 0.89 & 0.94 & 0.95 & 1.09 \\
\bottomrule
\end{tabular*}}
\end{table}

\begin{figure}[t]
\centering
\includegraphics[width=\linewidth]{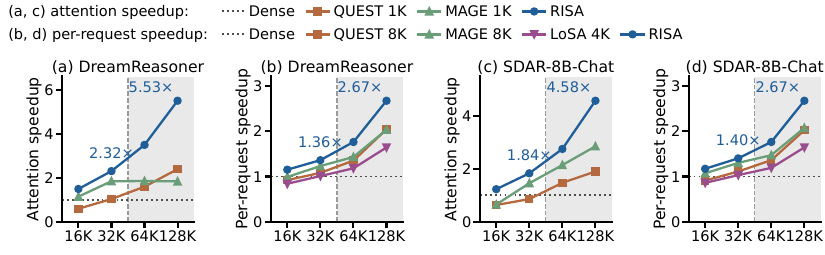}
\caption{\textbf{Efficiency across prefix lengths on both models.} (a,~c) Attention speedup over dense for DreamReasoner-8B and SDAR-8B-Chat (QUEST and MAGE at 1,024 tokens), from kernel profiles of 512-token requests with per-block work amortized over 15.6 or 10.3 steps per block. \revisiontext{(b,~d) Per-request speedup over dense, the values of \cref{tab:request}: complete 512-token decodes on dense's fixed trajectory (MAGE and QUEST at 8,192 tokens, LoSA at 4,096).} Dotted: dense ($1\times$). \revisiontext{Gray: beyond the configured 32K window, execution only.}}
\label{fig:attention_latency}\label{fig:step_speedup}\label{fig:efficiency}
\end{figure}

\paragraph{\revisiontext{Attention latency.}}
\revisiontext{\Cref{fig:frontier} sweeps the budget of every method and plots attention speedup against KL to dense within each model and context; at the default budgets, \method has the highest attention speedup in all eight in-window settings.} Attention latency covers GPU work for attention, selection and indices, normalization/merge, and required KV-cache operations, including attention in full-prefix finalization. Per-block work uses a common dense-derived amortization horizon of 15.6 steps per block for DreamReasoner and 10.3 for SDAR. \method has the lowest measured attention cost among the evaluated implementations at every length on both models (\cref{fig:efficiency}). On DreamReasoner, speedup over dense increases from $1.50\times$ at 16K to $2.32\times$, $3.51\times$, and $5.53\times$ at 32K, 64K, and 128K. Over this eightfold increase in prefix length, dense cost rises from $3.201$ to $18.763$\,ms, whereas \method rises from $2.135$ to $3.394$\,ms. The cap constrains repeated eligible-HOT reads, but seed, metadata, and finalization work still scale with history. The result is slower cost growth, not constant-time attention. This widening gap reflects the bounded read plan: as the prefix grows, dense attention repeatedly revisits more history, whereas \method limits the historical support consumed by later denoising steps. The same trend holds on SDAR: \method reaches $1.24\times$, $1.84\times$, $2.76\times$, and $4.58\times$ speedup as the prefix grows.

\paragraph{From attention savings to model execution.}
Attention savings matter only if they survive the rest of the forward pass. \revisiontext{\Cref{tab:request} times complete 512-token decodes on dense's fixed trajectory at each method's default budget. Per-request speedup grows with context on DreamReasoner and SDAR; on LLaDA2.2-mini, where attention is a smaller share of each step, it is smaller but present at every context.} \method is also faster \revisiontext{per request} than \revisiontext{every} sparse baseline at each measured length. \revisiontext{Across budgets on the same trajectory, each tested MAGE, QUEST, and LoSA budget has a faster \method point with no higher KL (\cref{fig:frontier_request,tab:oppoints}). When each method follows its own trajectory on book-prefixed MATH-500 at 32K, \method has the lowest profiled attention latency and mean per-request latency on DreamReasoner and SDAR and matches dense accuracy on the same 48 problems per model (\cref{tab:natural}); there, per-request speedup reflects both attention savings and changes in decoding length.} Projection, feed-forward, decoding, and synchronization remain outside the sparse-attention savings and determine how much of the attention-path improvement appears in \revisiontext{per-request} latency.

\begin{figure}[t]
\centering
\includegraphics[width=\linewidth]{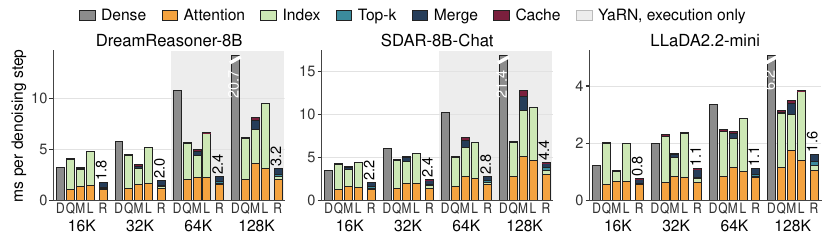}
\caption{\textbf{Attention latency breakdown.} \revisiontext{Attention-path kernel time per denoising step on an H200 NVL for a 512-token decode on a common prompt per model and prefix. Block work is amortized over dense's 22.2, 12.3, and 15.2 steps for DreamReasoner, SDAR, and LLaDA2.2-mini. D/Q/M/L/R denote Dense, QUEST, MAGE, LoSA, and \method. QUEST and MAGE use 8,192 tokens and LoSA uses 4,096. Index includes baseline selection and Top-k is \method's selector. Dense is the total. Gray marks execution with YaRN.}}
\label{fig:attention_composition}
\end{figure}

\paragraph{Where the attention path spends time.}
\revisiontext{\Cref{fig:attention_composition} separates prefix attention from Index, which covers scoring and baseline selection, and Top-k, the fused \method selector. Merge combines attention partials and observations, and Cache covers KV movement and layout maintenance. MAGE constructs its scores from the all-mask block observation, an expense that grows with the prefix even at a fixed selection budget. QUEST scores each query and attends the resulting union per KV head. LoSA reuses outputs for queries that change little, and its Index time exceeds its Attention time in all twelve settings. \method scores page summaries and reuses the bounded read plan across steps. It has the lowest summed kernel time among sparse methods at every model and prefix, with LoSA's totals 2.1--3.0$\times$ those of \method. The per-request measurements capture how these savings translate to full model execution.}

\paragraph{Scalability to Blackwell GPUs.}
We additionally measure amortized attention-path latency for SDAR-8B on an NVIDIA RTX PRO 6000 Blackwell Server Edition at a 64K-token prefix. Using SDAR-8B's measured average of 10.30 denoising steps per block, dense requires $10.43$\,ms per step, compared with $3.48$\,ms for \method ($2.99\times$ speedup), $7.18$\,ms for MAGE ($1.45\times$), and $5.39$\,ms for QUEST ($1.94\times$). The attention-path advantage therefore persists on a newer GPU generation.

\paragraph{Memory footprint.}
Recoverability trades storage for fewer reads. Peak allocation is $29.38$ versus dense's $27.62$\,GiB at 16K and $44.20$ versus $39.27$\,GiB at 64K. Packed KV, metadata, and scratch add storage in exchange for fewer reads during denoising.

\subsection{Ablation Study}
\label{sec:history}\label{sec:controls}\label{sec:ablations}
\begin{table}[t]
\centering\footnotesize
\caption{\textbf{Ablations on DreamReasoner and SDAR.} Each variant changes one component of the final configuration. \revisiontext{The workload uses a 16K prompt and 4,096 output tokens. Attention latency includes selection and amortized block work, profiled on one document per model (dense: 4.38 and 3.86\,ms). KV reads are relative to dense and, with KL to dense, are measured on six documents using dense's trajectories.}}
\label{tab:main_ablation}\label{tab:controls}
\setlength{\tabcolsep}{4pt}
{\revisioncolor
\begin{tabular*}{\linewidth}{@{\extracolsep{\fill}}lrrrrrr@{}}\toprule
 & \multicolumn{3}{c}{DreamReasoner} & \multicolumn{3}{c}{SDAR}\\\cmidrule(lr){2-4}\cmidrule(l){5-7}
Variant & \shortstack[r]{Attn.\\(ms/step)} & \shortstack[r]{KV reads\\(\%)} & \shortstack[r]{KL\\($10^{-3}$)} & \shortstack[r]{Attn.\\(ms/step)} & \shortstack[r]{KV reads\\(\%)} & \shortstack[r]{KL\\($10^{-3}$)}\\\midrule
\multicolumn{7}{@{}l}{\textit{Recovery and tiering}}\\
\quad No COLD recovery & 2.54 & 56 & 46.7 & 1.68 & 48 & 44.3 \\
\quad COLD reads spent on HOT & 2.62 & 59 & 45.1 & 1.75 & 56 & 45.1 \\
\quad Flat retrieval & 2.92 & 60 & 5.21 & 1.77 & 54 & 2.60 \\
\quad COLD quota of 16 pages & 2.65 & 57 & 14.5 & 1.71 & 49 & 13.9 \\
\midrule
\multicolumn{7}{@{}l}{\textit{Admission}}\\
\quad Recency admission & 2.83 & 60 & 4.10 & 1.76 & 54 & 1.70 \\
\quad Random admission & 2.74 & 60 & 4.14 & 1.76 & 54 & 1.72 \\
\quad Mass + learned correction & 2.81 & 60 & 4.08 & 1.85 & 54 & 1.76 \\
\midrule
\multicolumn{7}{@{}l}{\textit{Plan reuse and HOT cap}}\\
\quad Per-step reselection & 3.69 & 60 & 3.33 & 3.14 & 54 & 1.66 \\
\quad No HOT cap & 2.85 & 60 & 4.20 & 2.16 & 98 & 0.52 \\
\midrule
\textbf{\method (final)} & \textbf{2.86} & \textbf{60} & \textbf{4.21} & \textbf{1.77} & \textbf{54} & \textbf{1.74} \\
\bottomrule
\end{tabular*}}
\end{table}

\Cref{tab:main_ablation} isolates recoverable history, block-static plan reuse, and bounded HOT reads\revisiontext{, changing one component at a time on a workload in which generated history ages into COLD. Spending the recovery quota on HOT pages raises KL by more than an order of magnitude at comparable reads. Disabling recovery also raises KL. Separate tiers also beat a single ranking of all history at nearly identical reads, and a smaller COLD quota saves little time but raises KL. Plan reuse is what keeps attention cheap: rebuilding the plan at every step reads the same history and gains a little fidelity, but takes longer. On SDAR, lifting the HOT cap reads nearly the whole prefix and costs more attention time; on DreamReasoner, removing the cap barely changed realized reads in this workload.} These ablations separate the roles of the three components: recovery preserves access to older history, plan reuse amortizes support construction, and the HOT cap bounds recurring reads.

\revisiontext{Admission policies yield similar KL on the 16K ablation. On AIME24, where more generated history ages out of the grace window, mass-only admission lowers KL relative to recency and random admission at similar read volumes. It also approaches the fidelity of the learned correction (\cref{tab:admission}), which motivates using the mass score directly.}

\revisiontext{On LLaDA2.2-mini, COLD recovery lowers KL at comparable reads on both the 16K and 64K prompts (\cref{fig:history}). The needle test probes a different capability: retrieving distant evidence from the prompt while the HOT cap binds. From 32K to 128K, recall stays within six percentage points of dense while \method reads 32--48\% as much history. At 128K it matches dense recall while reading about a third as much.}

\revisiontext{Lowering the read fraction reduces realized reads at 16K on DreamReasoner. At 32K, the same reduction leaves the read volume nearly constant because both settings propose more eligible pages than the cap allows. Lowering the cap itself reduces reads further and raises KL. The read fraction thus controls the budget at shorter contexts, while the HOT cap sets the limit as the eligible history grows (\cref{app:tasks}).}

\paragraph{Block size.}
\revisiontext{With 64-token DreamReasoner blocks, \method still lowers both attention and per-request latency at 16K and 32K (\cref{tab:blocksize}). Each block size uses its own dense trajectory, and we hold the page size, grace window, and admission reservoir fixed in tokens. Changing the block size can change how many denoising steps share a plan, so we measure latency and KL against dense at the same block size. The cap binds at 32K for both block sizes, where \method retains the latency benefit while reading roughly 30\% of history. The measured speedup is lower at 64 tokens, even as the read ratio remains close at each prefix. At 32K, the 64-token configuration cuts attention time by more than half and still speeds up the complete request.}

\section{Conclusion}
\label{sec:conclusion}
\revisiontext{\method reduces repeated KV reads in block diffusion by keeping exact history recoverable and reusing a bounded read plan across denoising steps. Across eight in-window settings, it decodes faster than the tested MAGE, QUEST, and LoSA configurations at matched KL. On book-prefixed MATH-500 at 32K, it lowers per-request latency under native decoding on DreamReasoner and SDAR and matches dense answer accuracy over 48 paired problems per model.} The 128K H200 scaling results and the 64K SDAR Blackwell study show that the attention-path advantage persists as context and hardware change. \revisiontext{Keeping exact history resident trades KV-cache storage for fewer repeated reads.}

\subsection*{AI use statement}
Generative AI tools were used to edit author-written text for
grammar, clarity, and readability. The authors reviewed all AI-assisted
edits and take full responsibility for the final content of the paper.

\subsection*{Reproducibility statement}
The paper specifies the inference policy, decoding settings, and measurement scope. Static figures and directly editable tables accompany the source. Primary and supplementary studies use the measurement definitions and benchmark protocols documented in the paper and source.

\appendix
\label{appendix:start}
\raggedbottom
\setlength{\floatsep}{12pt plus 2pt minus 2pt}
\setlength{\textfloatsep}{14pt plus 2pt minus 2pt}
\setlength{\intextsep}{12pt plus 2pt minus 2pt}

\section{Protocol and Metrics}
\label{app:protocol}
\paragraph{Fidelity.} \revisiontext{Dense decodes each request with its native decoder. At every denoising step the method under test receives exactly dense's state (committed tokens and remaining masks), and we record $\mathrm{KL}(p_{\rm dense}\,\|\,p_{\rm method})$ at every masked position. We use KL rather than excess negative log-likelihood, which is not monotone in the number of reads.}
\paragraph{Latency.} \revisiontext{We report two latency metrics. \emph{Attention latency} is the time of the attention path per denoising step, in ms: attention, selection, indices, merging, required KV-cache work, and attention-related block work (seed step, planning, MAGE's all-mask scoring, finalization) spread over dense's steps per block, averaged over every step of the request. It is timed per request as the method's step time minus dense's non-attention time, with dense's attention time from kernel profiles; kernel profiles also apportion it into components (\cref{fig:attention_composition}), counting kernel time only. \emph{Per-request latency} is the measured wall time of a complete decode after prefill, including seed steps, planning, denoising, and finalization; we report it as total dense time over total method time across documents. Under the \emph{fixed trajectory} every method replays dense's denoising schedule on dense's states, so all methods run the same blocks and steps; under \emph{native decoding} (\cref{tab:free,tab:natural,tab:sens}) each method follows its own trajectory. Fixed-length timing uses 512 output tokens with end-of-sequence disabled, one request per GPU, and CUDA graphs. Timing excludes the warm-up document when applicable. The KV read ratio counts every historical KV row read, including protected, grace-window, and recovered pages, relative to dense.}
\paragraph{Context and block lengths.} \revisiontext{DreamReasoner-8B and SDAR-8B-Chat are configured for 32,768 positions. The 32K timing setting reserves 512 positions for decoding; their 64K and 128K settings use YaRN and measure execution only. The primary evaluation uses 32-token blocks, a reported operating point for DreamReasoner-8B and the native block size for SDAR-8B-Chat-b32 and LLaDA2.2-mini. \Cref{tab:blocksize} studies 32- and 64-token blocks on DreamReasoner.}

\section{Admission and History Selection}
\label{app:admission}
\paragraph{Mass score.} \revisiontext{The priority of a generated token $i$, computed separately for each layer and KV head, is the attention mass it received during its grace window:}
\begin{equation}
 \revisioncolor
 \pi_i=\log\bigl(1+w_i\bigr),\qquad w_i=\sum_{b\in\mathcal{G}(i)}\ \sum_{t}\ \sum_{r\in b,\ g\in h} p^{(t)}_{r,g}(i),
 \label{eq:priority}
\end{equation}
\revisiontext{summed over the blocks $\mathcal{G}(i)$ that read $i$ in the window, their denoising steps $t$, query rows $r$, and the query heads $g$ mapped to KV head $h$. Because $p$ is the probability of the sparse attention that already ran, scoring needs no extra forward pass.}
\paragraph{Allocator.} \revisiontext{When a source block leaves the grace window, each layer and KV head scores its 32 tokens with \cref{eq:priority}. Without score history it admits the top eight. Otherwise it admits the scores above the $1-8/32$ quantile of the previous $\min(n,24)\times 32$ scores, keeping between two and 24 tokens, and all 32 scores join the reservoir. Admitted exact KV enters HOT and the rest enters COLD; both remain stored. The last partial prompt block stays in HOT.}
\paragraph{Learned correction.} \revisiontext{The learned admission variant adds a bounded term, $\pi_i=\log(1+w_i)+4\tanh(\mathrm{MLP}(x_i,e_\ell))$, with six observation features $x_i$ (grace-window mass, mass at the admitting block, step-zero mass, top-eight hit count, key norm, and position in the block) and an eight-dimensional layer embedding $e_\ell$, 801 parameters in total. Training used 10,332 block--layer samples from 12 development prompts. The sparse model visits the states and a frozen dense teacher provides the target, the log of the mass a token receives from two blocks later onward.}
\paragraph{Admission on AIME24.} \revisiontext{At nearly identical read volumes, mass-only admission approaches the fidelity of the learned correction and improves on recency and random admission (\cref{tab:admission}). Ranking HOT and COLD jointly also increases KL at similar reads.}

\begin{table}[t]
\centering\small
\caption{\textbf{Admission and retrieval on AIME24.} Seven DreamReasoner problems with a 16,384-token output cap. Each variant uses a 128-page COLD quota. \revisiontext{KL to dense is measured on dense's denoising states.} Reads are realized historical KV reads relative to dense.}
\label{tab:admission}
{\revisioncolor
\begin{tabular}{@{}lrr@{}}\toprule
Rule & KL ($\times10^{-3}$) & KV reads (\%)\\\midrule
Mass + learned correction & 8.60 & 75.9 \\
Recency admission & 10.08 & 76.2 \\
Random admission & 10.30 & 76.0 \\
Flat retrieval & 11.15 & 76.6 \\
\midrule
\textbf{Mass only (\method)} & 8.72 & 76.0 \\
\bottomrule
\end{tabular}}
\end{table}

\paragraph{COLD quota.} \revisiontext{The default COLD quota is 128 pages per layer and KV head. \Cref{tab:main_ablation} compares it with a quota of 16 pages.}

\paragraph{Context-adaptive HOT cap (\method-K).} \revisiontext{\method-K sets the HOT cap to $\max(512,\lceil f|E_b|\rceil)$ pages per layer and KV head, where $E_b$ is the eligible HOT set. On tuning books, we choose the smallest tested $f$ whose KL at 128K is at most the fixed cap's KL at 32K. This selects $f=0.5$ (KL $0.0061$ against a target of $0.0085$). On twelve LLaDA2.2-mini evaluation documents, KL for the fixed and adaptive caps is $0.019$ and $0.012$ at 32K, $0.023$ and $0.008$ at 64K, and $0.055$ and $0.012$ at 128K on dense trajectories. On these fixed trajectories, \method-K is $1.04$--$1.17\times$ faster than dense per request.}

\section{LLaDA2.2-mini Adaptation}
\label{app:llada}
\revisiontext{Only the attention core is replaced: the checkpoint's fused projection, query/key normalization, rotary embedding, and output projection run unchanged, and dense, \method, and every baseline use the same attention kernel. The mixture-of-experts layers run as grouped matrix multiplications over the same routed experts, which reproduce the checkpoint's expert loop exactly and allow CUDA-graph capture. Tokens are committed by the checkpoint's confidence threshold with its token-editing step disabled for every method, so blocks keep 32 positions and finished blocks are immutable. BF16 rounding alone changes expert routing in this model, so every method's KL includes a small numerical floor.}

\section{Additional Efficiency Results}
\label{app:scaling}
\begin{figure}[H]
\centering
\includegraphics[width=\linewidth]{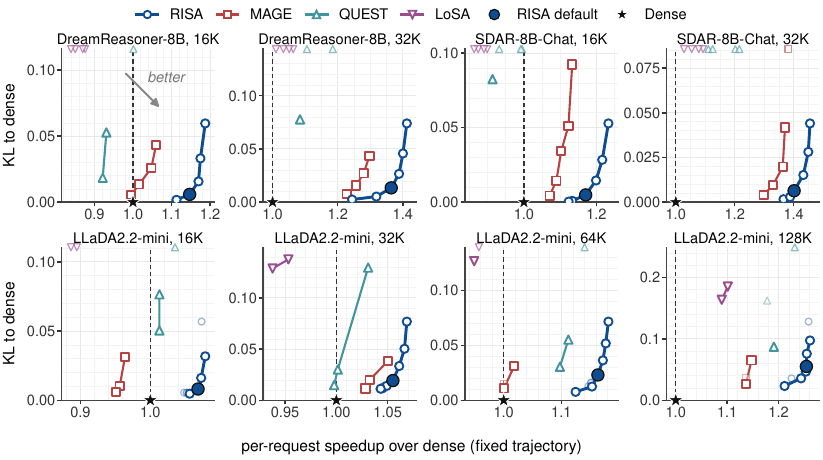}
\vspace{-6pt}
\caption{\revisiontext{\textbf{Quality--latency Pareto frontiers for complete requests.} Every budget of every method, placed by per-request speedup over dense and KL to dense; lower right is better. Speedup is total dense wall time over total method wall time for complete 512-token decodes replayed on dense's denoising trajectory, and KL is measured on the same requests; the star at $(1, 0)$ is dense. Lines join each method's Pareto budgets within the panel (no other budget of the same method is both faster and lower in KL), faded markers are budgets beaten by another budget of the same method, and the filled circle is \method's default. Points whose KL exceeds a panel's range are drawn on its top edge at their measured speedup.}}
\label{fig:frontier_request}
\end{figure}

\revisiontext{\Cref{tab:perf} reports attention-latency and per-request speedup for every model, context, and method, \cref{fig:frontier_request} repeats the budget sweep of \cref{fig:frontier} with per-request latency on the fixed trajectory, and \cref{tab:free} reports per-request speedup under native decoding, where each method follows its own decoding trajectory.}

\begin{figure}[H]
\centering
\includegraphics[width=\linewidth]{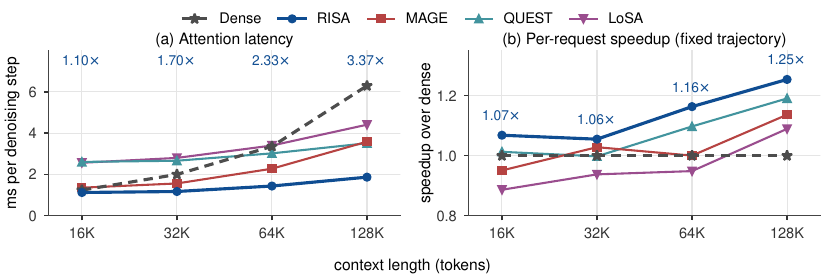}
\caption{\revisiontext{\textbf{LLaDA2.2-mini at 16K--128K}. (a) Attention latency in ms per denoising step for dense (dashed) and each method, measured as in \cref{tab:perf} on 512-token requests with per-block work spread over dense's steps per block; \method's attention-latency speedup over dense is printed above each length. (b) Per-request speedup over dense (dashed, $1\times$) for complete 512-token requests replayed on dense's denoising trajectory. \method's default, MAGE and QUEST at 8,192 tokens, and LoSA at 4,096 tokens.}}
\label{fig:llada}\label{fig:latency}
\end{figure}

\revisiontext{\Cref{fig:llada} reports both latency metrics on LLaDA2.2-mini: attention latency (a), measured as in \cref{tab:perf}, and per-request speedup on the fixed trajectory (b). The per-request speedup is smaller because attention is a modest share of each step on this mixture-of-experts model. The kernel-profile components in \cref{fig:attention_composition} count kernel time only, so their totals are lower than the timed attention latency. \method is the fastest method on both metrics at every length.}

\begin{table}[H]
\centering\scriptsize
\caption{\revisiontext{\textbf{Attention-latency and per-request speedup over dense} at the default budgets (\method's default, MAGE and QUEST at 8,192 tokens, LoSA at 4,096), for every measured model, context, and method. Attention speedup is dense's attention latency (ms per denoising step; the Dense attn.\ column) over the method's, timed on 512-token requests in which each method decodes natively, with each method's per-block work (seed step, planning, MAGE's all-mask scoring, finalization) spread over dense's steps per block in the same session. Per-request speedup is total dense wall time over total method wall time for complete 512-token requests replayed on dense's denoising trajectory. Read is \method's KV read ratio. Bold marks the fastest method. $^\dagger$Beyond the configured window (YaRN); execution only.}}
\label{tab:perf}
\setlength{\tabcolsep}{2.8pt}
{\revisioncolor
\begin{tabular*}{\linewidth}{@{\extracolsep{\fill}}llrrrrrrrrrr@{}}\toprule
 & & Dense attn. & \multicolumn{4}{c}{Attention-latency speedup} & \multicolumn{4}{c}{Per-request speedup (fixed trajectory)} & \\\cmidrule(lr){4-7}\cmidrule(lr){8-11}
Model & Context & (ms/step) & \method & MAGE 8K & QUEST 8K & LoSA 4K & \method & MAGE 8K & QUEST 8K & LoSA 4K & Read\\\midrule
DreamReasoner-8B & 16K & 3.27 & \textbf{1.59} & 1.10 & 0.67 & 0.62 & \textbf{1.15} & 0.99 & 0.92 & 0.84 & 0.54 \\
 & 32K & 5.95 & \textbf{2.67} & 1.62 & 1.15 & 1.02 & \textbf{1.36} & 1.23 & 1.08 & 1.01 & 0.31 \\
 & 64K$^\dagger$ & 9.90 & \textbf{3.95} & 2.30 & 1.77 & 1.43 & \textbf{1.76} & 1.43 & 1.35 & 1.18 & 0.17 \\
 & 128K$^\dagger$ & 20.31 & \textbf{8.88} & 3.49 & 3.52 & 2.27 & \textbf{2.67} & 2.03 & 2.04 & 1.64 & 0.10 \\
\addlinespace[2pt]
SDAR-8B-Chat & 16K & 3.17 & \textbf{1.72} & 1.38 & 0.69 & 0.64 & \textbf{1.17} & 1.07 & 0.91 & 0.86 & 0.54 \\
 & 32K & 5.79 & \textbf{3.02} & 2.09 & 1.20 & 1.05 & \textbf{1.40} & 1.30 & 1.11 & 1.03 & 0.29 \\
 & 64K$^\dagger$ & 9.74 & \textbf{4.16} & 2.51 & 1.79 & 1.44 & \textbf{1.76} & 1.47 & 1.36 & 1.19 & 0.16 \\
 & 128K$^\dagger$ & 20.42 & \textbf{8.63} & 3.70 & 3.47 & 2.27 & \textbf{2.67} & 2.07 & 2.02 & 1.64 & 0.10 \\
\addlinespace[2pt]
LLaDA2.2-mini & 16K & 1.24 & \textbf{1.10} & 0.91 & 0.48 & 0.48 & \textbf{1.07} & 0.95 & 1.01 & 0.89 & 0.56 \\
 & 32K & 2.00 & \textbf{1.70} & 1.28 & 0.75 & 0.71 & \textbf{1.06} & 1.03 & 1.00 & 0.94 & 0.34 \\
 & 64K & 3.36 & \textbf{2.33} & 1.47 & 1.11 & 0.99 & \textbf{1.16} & 1.00 & 1.10 & 0.95 & 0.23 \\
 & 128K & 6.29 & \textbf{3.37} & 1.76 & 1.79 & 1.44 & \textbf{1.25} & 1.14 & 1.19 & 1.09 & 0.17 \\
\bottomrule
\end{tabular*}}
\end{table}

\begin{table}[H]
\centering\scriptsize
\caption{\revisiontext{\textbf{Per-request speedup under native decoding} (decode wall time; dense $=1.00\times$). Each method, dense included, decodes its own 512-token trajectory on the same documents (DreamReasoner: 30 at 16K and 32K, six at 64K and 128K; SDAR: six; LLaDA2.2-mini: 12); speedup is total dense time over total method time. \method's default, MAGE and QUEST at 8,192 tokens, LoSA at 4,096. Bold: fastest sparse method per column. $^\dagger$Beyond DreamReasoner's and SDAR's configured 32K window (YaRN); execution only.}}
\label{tab:free}
\setlength{\tabcolsep}{2.6pt}
{\revisioncolor
\begin{tabular*}{\linewidth}{@{\extracolsep{\fill}}lrrrrrrrrrrrr@{}}\toprule
 & \multicolumn{4}{c}{DreamReasoner-8B} & \multicolumn{4}{c}{SDAR-8B-Chat} & \multicolumn{4}{c}{LLaDA2.2-mini}\\\cmidrule(lr){2-5}\cmidrule(lr){6-9}\cmidrule(l){10-13}
Method & 16K & 32K & 64K$^\dagger$ & 128K$^\dagger$ & 16K & 32K & 64K$^\dagger$ & 128K$^\dagger$ & 16K & 32K & 64K & 128K\\\midrule
\method & \textbf{1.20} & \textbf{1.51} & \textbf{2.10} & \textbf{3.05} & \textbf{1.16} & \textbf{1.47} & \textbf{1.87} & \textbf{2.96} & \textbf{0.98} & \textbf{1.12} & \textbf{1.11} & \textbf{1.20} \\
MAGE 8K & 0.89 & 1.26 & 1.54 & 2.17 & 1.04 & 1.32 & 1.51 & 2.20 & 0.77 & 1.04 & 0.94 & 1.06 \\
QUEST 8K & 0.79 & 1.01 & 1.39 & 2.24 & 0.88 & 1.08 & 1.39 & 2.17 & 0.86 & 1.06 & 1.01 & 1.03 \\
LoSA 4K & 0.72 & 0.95 & 1.14 & 1.62 & 0.80 & 0.99 & 1.20 & 1.71 & 0.75 & 0.78 & 0.86 & 1.00 \\
\bottomrule
\end{tabular*}}
\end{table}

\paragraph{Absolute latency at matched fidelity.} \revisiontext{\Cref{tab:oppoints} reports measured seconds per request at 32K for every MAGE, QUEST, and LoSA budget and, for each, the fastest \method budget with no higher KL, on paired documents with dense's fixed trajectory. Each of the 28 baseline operating points has a faster \method counterpart with no higher KL.}

\begin{table}[H]
\centering\scriptsize
\caption{\revisiontext{\textbf{Absolute per-request latency at matched fidelity, 32K.} Every method follows dense's fixed denoising trajectory on the same documents. Seconds are decode wall time for 512 tokens, excluding the warm-up document. KL ($\times10^{-3}$) and the KV read ratio include all paired documents. For each baseline budget, the matched \method budget is the fastest one whose KL is not higher (HOT cap in tokens, read fraction $\rho$). Dense's seconds per request are given beside each model.}}
\label{tab:oppoints}
\setlength{\tabcolsep}{2.6pt}
{\revisioncolor
\begin{tabular}{@{}llrrrlrrr@{}}\toprule
 & \multicolumn{4}{c}{Baseline} & \multicolumn{4}{c}{Matched \method}\\\cmidrule(lr){2-5}\cmidrule(l){6-9}
Model (dense s) & Budget & s & KL & Read & Budget & s & KL & Read\\\midrule
DreamReasoner-8B (4.09 s) & MAGE 1,024 & 3.18 & 43.9 & 0.14 & 4,096, $\rho$=0.5 & 2.95 & 28.4 & 0.19 \\
 & MAGE 2,048 & 3.22 & 26.6 & 0.17 & 8,192, $\rho$=0.5 & 3.00 & 14.4 & 0.31 \\
 & MAGE 4,096 & 3.28 & 14.2 & 0.23 & 16,384, $\rho$=1.0 & 3.10 & 5.6 & 0.55 \\
 & MAGE 8,192 & 3.39 & 6.8 & 0.34 & 16,384, $\rho$=1.0 & 3.10 & 5.6 & 0.55 \\
 & QUEST 1,024 & 3.44 & 326.5 & 0.09 & 1,024, $\rho$=0.5 & 2.91 & 79.3 & 0.10 \\
 & QUEST 4,096 & 3.70 & 125.5 & 0.18 & 1,024, $\rho$=0.5 & 2.91 & 79.3 & 0.10 \\
 & QUEST 8,192 & 3.76 & 72.6 & 0.29 & 2,048, $\rho$=0.5 & 2.93 & 49.5 & 0.13 \\
 & LoSA 1,024 & 3.92 & 149.0 & 0.14 & 1,024, $\rho$=0.5 & 2.91 & 79.3 & 0.10 \\
 & LoSA 4,096 & 4.06 & 140.1 & 0.34 & 1,024, $\rho$=0.5 & 2.91 & 79.3 & 0.10 \\
\addlinespace[2pt]
SDAR-8B-Chat (7.13 s) & MAGE 512 & 5.13 & 80.2 & 0.10 & 1,024, $\rho$=1.0 & 4.82 & 44.4 & 0.08 \\
 & MAGE 1,024 & 5.18 & 41.8 & 0.11 & 2,048, $\rho$=1.0 & 4.85 & 28.0 & 0.11 \\
 & MAGE 2,048 & 5.22 & 19.8 & 0.14 & 4,096, $\rho$=1.0 & 4.93 & 15.3 & 0.17 \\
 & MAGE 4,096 & 5.31 & 9.5 & 0.20 & 8,192, $\rho$=1.0 & 5.02 & 6.5 & 0.29 \\
 & MAGE 8,192 & 5.42 & 4.0 & 0.32 & 12,288, $\rho$=1.0 & 5.12 & 3.3 & 0.41 \\
 & QUEST 1,024 & 5.88 & 351.7 & 0.09 & 1,024, $\rho$=1.0 & 4.82 & 44.4 & 0.08 \\
 & QUEST 2,048 & 5.95 & 218.0 & 0.12 & 1,024, $\rho$=1.0 & 4.82 & 44.4 & 0.08 \\
 & QUEST 4,096 & 6.34 & 149.1 & 0.17 & 1,024, $\rho$=1.0 & 4.82 & 44.4 & 0.08 \\
 & QUEST 8,192 & 6.47 & 93.2 & 0.29 & 1,024, $\rho$=1.0 & 4.82 & 44.4 & 0.08 \\
 & LoSA 1,024 & 6.65 & 161.1 & 0.12 & 1,024, $\rho$=1.0 & 4.82 & 44.4 & 0.08 \\
 & LoSA 4,096 & 6.90 & 156.5 & 0.32 & 1,024, $\rho$=1.0 & 4.82 & 44.4 & 0.08 \\
\addlinespace[2pt]
LLaDA2.2-mini (2.95 s) & MAGE 1,024 & 2.84 & 38.3 & 0.22 & 4,096, $\rho$=0.5 & 2.79 & 33.5 & 0.23 \\
 & MAGE 4,096 & 2.87 & 20.1 & 0.30 & 8,192, $\rho$=0.5 & 2.81 & 19.3 & 0.34 \\
 & MAGE 8,192 & 2.88 & 11.5 & 0.40 & 16,384, $\rho$=0.75 & 2.85 & 11.3 & 0.57 \\
 & QUEST 1,024 & 2.88 & 129.5 & 0.13 & 1,024, $\rho$=0.5 & 2.78 & 76.8 & 0.14 \\
 & QUEST 4,096 & 2.95 & 30.4 & 0.21 & 8,192, $\rho$=0.5 & 2.81 & 19.3 & 0.34 \\
 & QUEST 8,192 & 2.97 & 15.0 & 0.33 & 12,288, $\rho$=0.75 & 2.83 & 14.1 & 0.45 \\
 & LoSA 1,024 & 3.10 & 137.5 & 0.19 & 1,024, $\rho$=0.5 & 2.78 & 76.8 & 0.14 \\
 & LoSA 4,096 & 3.15 & 128.6 & 0.37 & 1,024, $\rho$=0.5 & 2.78 & 76.8 & 0.14 \\
\bottomrule
\end{tabular}}
\end{table}

\paragraph{Block size.} \revisiontext{\method reduces both attention and request latency at 32 and 64 tokens on DreamReasoner-8B (\cref{tab:blocksize}). At 64 tokens the request speedup is $1.09\times$ at 16K and $1.28\times$ at 32K, with the HOT cap active in every measured 32K block. The model was trained and evaluated at block sizes 4, 8, 16, and 32 \citep{wu2026dreamreasoner}, so 64 tokens is an execution stress test. Each measured block size uses its own dense trajectory and therefore its own KL reference.}

\begin{table}[H]
\centering\scriptsize
\caption{\revisiontext{\textbf{Block-size sweep on DreamReasoner-8B.} An execution test of dense and \method only, on DreamReasoner-8B at 16K and 32K with 12 documents per cell. DreamReasoner was trained and evaluated at blocks of 4--32 tokens, and 64-token blocks extend the range tested here. Request latency (decode wall time, s), KL, reads, and cap activity use 512-token decodes replayed on each block size's own dense trajectory, with KL referenced to the dense run at that block size. Attention profiles cover output positions 320--511 and amortize block work over dense's steps per block. Both block sizes use 16-token pages, an 8,192-token HOT cap, a 256-token grace window, a 128-page COLD quota, $\rho=0.5$, and mass-only admission; the allocator preserves the admission rate and the reservoir size in tokens. Cap activity is the share of \method's sparse blocks whose reused plan hits the HOT cap.}}
\label{tab:blocksize}
\setlength{\tabcolsep}{2.8pt}
{\revisioncolor
\begin{tabular}{@{}llrrrrrrrrr@{}}\toprule
 & & \multicolumn{3}{c}{Attention (ms / step)} & \multicolumn{3}{c}{Request (s)} & & & \\\cmidrule(lr){3-5}\cmidrule(lr){6-8}
Block & Context & Dense & \method & Speedup & Dense & \method & Speedup & KL ($\times10^{-3}$) & Reads (\%) & Cap (\%)\\\midrule
32 & 16K & 3.19 & 1.84 & 1.73$\times$ & 3.78 & 3.28 & 1.15$\times$ & 5.3 & 54 & 0 \\
 & 32K & 5.95 & 2.06 & 2.89$\times$ & 4.14 & 3.06 & 1.36$\times$ & 14.4 & 31 & 100 \\
\addlinespace[2pt]
64 & 16K & 3.36 & 2.60 & 1.30$\times$ & 4.19 & 3.84 & 1.09$\times$ & 6.9 & 52 & 0 \\
 & 32K & 5.81 & 2.76 & 2.11$\times$ & 4.31 & 3.38 & 1.28$\times$ & 16.1 & 29 & 100 \\
\bottomrule
\end{tabular}}
\end{table}

\paragraph{Complete requests on a natural long task.} \revisiontext{\Cref{tab:natural} evaluates native decoding on book-prefixed MATH-500 at 32K. Every method decodes the same 48 prompts per model. \method leads in attention and per-request speedup on DreamReasoner and SDAR and matches dense answer accuracy on both.}

\begin{table}[H]
\centering\scriptsize
\caption{\revisiontext{\textbf{Native decoding on book-prefixed MATH-500.} Each setting uses 48 paired problems and a 2,048-token output cap. Each method follows its own trajectory. Cells report attention and per-request speedup over dense (top), then mean decode seconds, answer accuracy (\%), and mean output tokens (bottom). Attention speedups use one profiled request per method and setting. MAGE and QUEST use 8,192 tokens and LoSA uses 4,096.}}
\label{tab:natural}
\setlength{\tabcolsep}{3pt}
{\revisioncolor
\begin{tabular*}{\linewidth}{@{\extracolsep{\fill}}lccccc@{}}\toprule
Model, context & Dense & \method & MAGE & QUEST & LoSA \\\midrule
DreamReasoner, 32K & \shortstack[c]{1.00 / 1.00\\5.8 / 77.1 / 795} & \shortstack[c]{2.69 / 1.55\\3.7 / 77.1 / 771} & \shortstack[c]{1.50 / 1.27\\4.6 / 79.2 / 778} & \shortstack[c]{1.36 / 1.16\\5.0 / 81.2 / 758} & \shortstack[c]{1.24 / 0.89\\6.5 / 75.0 / 882} \\
\addlinespace[2pt]
SDAR, 32K & \shortstack[c]{1.00 / 1.00\\6.9 / 62.5 / 587} & \shortstack[c]{3.02 / 1.47\\4.7 / 62.5 / 585} & \shortstack[c]{1.95 / 1.30\\5.3 / 62.5 / 604} & \shortstack[c]{1.37 / 1.07\\6.5 / 60.4 / 599} & \shortstack[c]{1.26 / 0.92\\7.5 / 56.2 / 657} \\
\bottomrule
\end{tabular*}}
\end{table}

\section{Additional Quality Results}
\label{app:tasks}
\revisiontext{We report one-factor sensitivity, multi-needle retrieval, and fidelity on benchmark prompts. On DreamReasoner at 16K, lowering $\rho$ from 0.5 to 0.25 nearly triples KL as fewer HOT pages are read. At 32K, the cap holds reads near 31\% over the same change in $\rho$. Lowering the cap itself to 2,048 tokens at 32K raises KL about fourfold (\cref{tab:sens}).}

\begin{table}[H]
\centering\scriptsize
\caption{\revisiontext{\textbf{One-factor sensitivity} around the default on three further documents. Each cell gives per-request speedup over dense under native decoding, where every method, dense included, decodes its own 512-token trajectory, followed by KL to dense measured separately on dense's denoising states for the same documents. The default $\rho$ is 0.5 on DreamReasoner and 1.0 on SDAR.}}
\label{tab:sens}
\setlength{\tabcolsep}{3pt}
{\revisioncolor
\begin{tabular}{@{}lrrrr@{}}\toprule
Variant & DreamReasoner 16K & DreamReasoner 32K & SDAR 16K & SDAR 32K\\\midrule
cap $K_{\max}$ = 128 pages (2,048 tokens) & 1.26$\times$ / 0.0292 & 1.52$\times$ / 0.0575 & 1.21$\times$ / 0.0138 & 1.51$\times$ / 0.0167 \\
cap $K_{\max}$ = 256 pages & 0.96$\times$ / 0.0127 & 1.46$\times$ / 0.0305 & 1.20$\times$ / 0.0054 & 1.50$\times$ / 0.0077 \\
cap $K_{\max}$ = 512 pages (default) & 1.14$\times$ / 0.0045 & 1.61$\times$ / 0.0135 & 1.17$\times$ / 0.0012 & 1.47$\times$ / 0.0028 \\
cap $K_{\max}$ = 1,024 pages & 1.14$\times$ / 0.0045 & 1.40$\times$ / 0.0052 & 1.12$\times$ / 0.0004 & 1.39$\times$ / 0.0008 \\
no cap & 1.14$\times$ / 0.0045 & 1.40$\times$ / 0.0052 & 1.12$\times$ / 0.0004 & 1.30$\times$ / 0.0004 \\
$\rho$ = 0.25 & 0.99$\times$ / 0.0132 & 1.74$\times$ / 0.0139 & 1.20$\times$ / 0.0054 & 1.47$\times$ / 0.0029 \\
$\rho$ = 0.5 & 1.14$\times$ / 0.0045 & 1.61$\times$ / 0.0135 & 1.18$\times$ / 0.0012 & 1.46$\times$ / 0.0028 \\
$\rho$ = 0.75 & 0.96$\times$ / 0.0046 & 1.59$\times$ / 0.0135 & 1.17$\times$ / 0.0012 & 1.47$\times$ / 0.0028 \\
$\rho$ = 1.0 & 0.97$\times$ / 0.0046 & 1.60$\times$ / 0.0135 & 1.17$\times$ / 0.0012 & 1.47$\times$ / 0.0028 \\
grace $G$ = 4 & 1.12$\times$ / 0.0045 & 1.59$\times$ / 0.0134 & 1.16$\times$ / 0.0012 & 1.46$\times$ / 0.0028 \\
per-step reselection & 0.89$\times$ / 0.0040 & 1.33$\times$ / 0.0105 & 0.98$\times$ / 0.0011 & 1.17$\times$ / 0.0027 \\
\bottomrule
\end{tabular}}
\end{table}

\begin{table}[H]
\centering\scriptsize
\caption{\revisiontext{\textbf{Multi-needle retrieval in long prompts.} Three needles with two distractors in 16 documents, two layouts ($n=32$ per cell). Each cell gives in-order exact match (all three needles answered in order) / per-needle recall (share of the three needles retrieved), in \%. The final rows give \method's realized KV reads relative to dense and the share of its sparse blocks whose reused plan hits the HOT cap, averaged over requests. MAGE and QUEST use 4,096 tokens and LoSA 1,024.}}
\label{tab:cf3}
\setlength{\tabcolsep}{3pt}
{\revisioncolor
\begin{tabular}{@{}lrrrr@{}}\toprule
Method & DreamReasoner 16K & DreamReasoner 32K & SDAR 16K & SDAR 32K\\\midrule
Dense & 46.9 / 92.7 & 43.8 / 93.8 & 6.2 / 51.0 & 18.8 / 68.8 \\
\method (default) & 46.9 / 92.7 & 40.6 / 92.7 & 6.2 / 51.0 & 12.5 / 65.6 \\
MAGE 4K & 43.8 / 92.7 & 40.6 / 91.7 & 6.2 / 47.9 & 9.4 / 58.3 \\
QUEST 4K & 46.9 / 94.8 & 34.4 / 84.4 & 6.2 / 42.7 & 6.2 / 56.2 \\
LoSA 1K & 31.2 / 80.2 & 31.2 / 81.2 & 18.8 / 64.6 & 34.4 / 78.1 \\
\midrule
\method KV reads (\%) & 53.8 & 29.7 & 53.1 & 29.1 \\
\method HOT cap active (\% of blocks) & 0 & 100 & 100 & 100 \\
\bottomrule
\end{tabular}}
\end{table}

\begin{figure}[H]
\centering
\includegraphics[width=\linewidth]{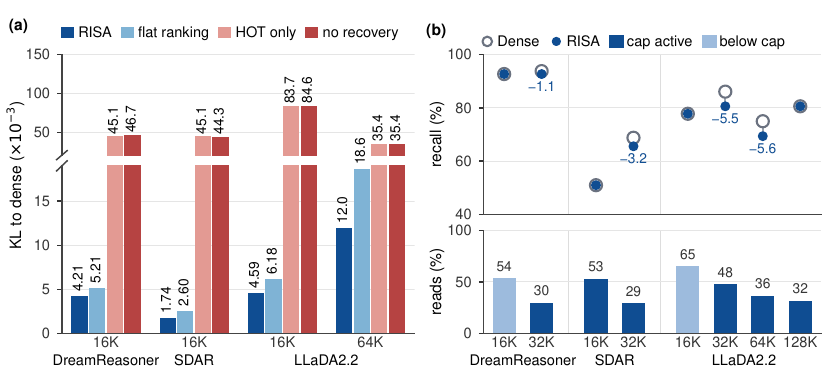}
\caption{\revisiontext{\textbf{Recovery and retrieval under bounded reads.} (a) Generated-history recovery: KL to dense ($\times10^{-3}$) after 4,096 generated tokens (six documents per model and context) for \method, flat retrieval (one ranking over HOT and COLD), HOT-only (recovery reads spent on HOT pages), and no recovery. The first three use comparable read volumes where measured. (b) Prompt retrieval of three needles among distractors ($n=32$; 12 on LLaDA2.2-mini): per-needle recall for dense and \method (top, with gaps in percentage points) and \method's KV reads relative to dense (bottom; dark: HOT cap active in at least half of sparse blocks).}}
\label{fig:history}
\end{figure}
\begin{table}[H]
\centering\scriptsize
\caption{\revisiontext{\textbf{Fidelity on benchmark prompts}: KL to dense along dense's own generation. In long-context MATH-500, a MATH-500 problem follows a long document, so generated history ages into COLD before the answer is complete.}}
\label{tab:tasks}
\setlength{\tabcolsep}{2.6pt}
{\revisioncolor
\begin{tabular}{@{}lrrrr@{}}\toprule
Method & HotpotQA & Long-context MATH-500 & MATH-500 & GSM8K\\
\midrule
\multicolumn{5}{@{}l}{\textit{DreamReasoner-8B}}\\
\method (default) & 0.0087 & 0.0090 & 0.0031 & 0.0013 \\
\method, no recovery ($q=0$) & 0.0141 & 0.0479 & 0.0232 & 0.0013 \\
MAGE 8K & 0.0055 & 0.0054 & 0.0003 & 0.0001 \\
QUEST 8K & 0.0071 & 0.0115 & 0.0005 & 0.0001 \\
LoSA 1K & 0.2101 & 0.1687 & 0.0456 & 0.0074 \\
\midrule
\multicolumn{5}{@{}l}{\textit{SDAR-8B-Chat}}\\
\method (default) & 0.0012 & 0.0034 & 0.0003 & 0.0000 \\
\method, no recovery ($q=0$) & 0.0012 & 0.0269 & 0.0003 & 0.0000 \\
MAGE 8K & 0.0010 & 0.0019 & 0.0002 & 0.0000 \\
QUEST 8K & 0.0049 & 0.0050 & 0.0003 & 0.0000 \\
LoSA 1K & 0.0396 & 0.1864 & 0.0346 & 0.0000 \\
\bottomrule
\end{tabular}}
\end{table}

\section{Attention Profiling Protocol}
\label{app:timing_intervals}
\begin{table}[H]
\centering\small
\caption{\textbf{Absolute efficiency results on DreamReasoner-8B.} \revisiontext{Attention latency in milliseconds per denoising step for the original configuration (QUEST and MAGE at 1,024 tokens), from one profiled source per context with block work amortized over 15.6 steps; 64K and 128K lie beyond the configured 32K window and measure execution only.} Bold marks the fastest sparse method.}
\label{tab:attention_latency}\label{tab:latency}
\begin{tabular}{@{}lrrrr@{}}\toprule
Method & 16K & 32K & 64K & 128K\\\midrule
Dense & 3.201 & 5.438 & 9.829 & 18.763\\\midrule
QUEST & 5.355 & 5.211 & 6.123 & 7.772\\
MAGE & 2.785 & 2.954 & 5.280 & 10.159\\
\method & \textbf{2.135} & \textbf{2.348} & \textbf{2.799} & \textbf{3.394}\\\bottomrule
\end{tabular}
\end{table}

\Cref{tab:attention_latency} records the absolute attention measurements underlying panels (a) and (c) of \cref{fig:efficiency}. Attention includes selection, indices, merge, cache operations, and finalization attention. Block work uses the common 15.6-step horizon on DreamReasoner and 10.3-step horizon on SDAR.

\paragraph{Measurement scope.}
Kernels are assigned to their serving call sites. Captured kernels inherit same-name stage shares from an eager run of the same requests, while fused work is counted once. For the attention scaling in \cref{fig:efficiency}, kernel profiles cover blocks 10--15 of a 512-token decode. \revisiontext{For \cref{fig:attention_composition}, we fit kernel time as a function of denoising-step count across all output blocks after the first and amortize the fitted per-block cost over the model's mean dense steps. Index totals the positive excess over dense in matrix multiplications, elementwise operations, and other kernels. Merge and Cache record the excess epilogue and memory operation time, respectively. The stacks sum kernel durations, including overlapping kernels such as asynchronous selection.}

\paragraph{Memory footprint.}
\label{app:memory_amortization}
Exact recovery does not imply KV compression. Peak allocation rises from dense's 27.62 to 29.38\,GiB at 16K and from 39.27 to 44.20\,GiB at 64K. Packed copies, metadata, and scratch trade additional storage for fewer repeated reads, consistent with \method's focus on read efficiency rather than cache compression.

\section{Block Execution Algorithm}
\label{app:algorithm}
The pseudocode below makes the ordering explicit: the seed forward observes the outgoing grace-window block before admission, the resulting seed queries build one historical read plan if denoising continues, and finalization remains a full-prefix forward pass. \revisiontext{\textsc{Admit} applies the mass score in \cref{eq:priority} and the allocator in \cref{app:admission}.}

\begin{algorithm}[H]
\footnotesize
\DontPrintSemicolon
\SetAlgoNlRelativeSize{-1}
\SetAlCapSkip{0.5ex}
\caption{\method: denoising one block.}
\label{alg:block}
\KwIn{Masked block $x$; KV cache $M$; statistics $A$;
      frozen model $F$}
\KwOut{Committed block $x$; updated $M,A$}
$(M,a)\gets\operatorname{StageRecent}(M,G)$\;
$(z,Q,o)\gets\operatorname{SeedForward}(F,x,M,q)$\;
$A\gets\operatorname{Update}(A,o)$\;
\lIf{$a\neq\varnothing$}{$(M,A)\gets\operatorname{Admit}(M,a,A)$}
$x\gets\operatorname{Commit}(x,z,0)$; $t\gets1$\;
\If{$\operatorname{Continue}(x,t)$}{
 \ForEach{layer $\ell$ and KV head $h$}{
  $(H,C,R)\gets\operatorname{Pages}(M,\ell,h)$\;
  $H_p\gets\operatorname{First}_4(H)\cup\operatorname{Last}_8(H)$;
  $E\gets H\setminus H_p$\;
  $u\gets\operatorname{Score}(Q_{\ell,h},E\cup C)$;
  $k\gets\operatorname{Budget}(|E|)$\;
  $S_{\ell,h}\gets R\cup\operatorname{rows}\bigl(H_p\cup\operatorname{Top}_k(u,E)\cup\operatorname{Top}_q(u,C)\bigr)$\;
 }
 \While{$\operatorname{Continue}(x,t)$}{
  $(z,o)\gets\operatorname{SparseForward}(F,x,M,S)$\;
  $A\gets\operatorname{Update}(A,o)$\;
  $x\gets\operatorname{Commit}(x,z,t)$; $t\gets t+1$\;
 }
}
$M\gets\operatorname{FinalizeKV}(F,x,M)$\;
\Return{$(x,M,A)$}\;
\end{algorithm}

\section{Additional Evaluation Details}
\label{app:quality_robustness}
\paragraph{Benchmark protocols.}
DreamReasoner-8B is evaluated on GSM8K, MATH-500, AIME25, LiveCodeBench v6, and HotpotQA; SDAR-8B-Chat is evaluated on GSM8K, MATH-500, and HotpotQA. All methods use identical prompts, tokenization, decoding settings, and generation limits within each model and benchmark. GSM8K and MATH use mathematical-answer scoring \citep{cobbe2021gsm8k,hendrycks2021math}; LiveCodeBench \citep{jain2024livecodebench} uses execution pass@1; HotpotQA uses LongBench token F1 \citep{bai2024longbench}. For LiveCodeBench, the final fenced Python program must pass the benchmark tests under Python 3.12.13; missing code scores zero. AIME25 uses temperature $0.6$ with a 32,768 total-position limit.

\paragraph{Configuration.}
The same frozen configuration is used across the reported tasks for each model. DreamReasoner uses $G=8$, $K_{\max}=512$, $q=16$, and $\rho=0.5$; SDAR uses the same history-management settings with \revisiontext{$\rho=0.75$}. \revisiontext{Both use the learned admission correction.} QUEST and MAGE use their stated 1,024-token budgets under the common carrier and decoder used throughout the main evaluation.

\paragraph{Motivation cohort.}
Motivation diagnostics use a fixed request cohort and sample every fourth eligible block at five layers and all KV heads. Recovery error is per-query relative Euclidean error against full-domain attention, with a $10^{-6}$ denominator floor. At quotas 16/32/64, additional HOT candidates are insufficient in 2.1/48.2/84.9\% of layer/head states; these mismatches are retained rather than treated as successful matches in the reported summaries.

\label{appendix:end}
\end{document}